\documentclass[conference]{IEEEtran}

\IEEEoverridecommandlockouts

\usepackage[switch]{lineno}

\AtBeginDocument{%
  }

\usepackage{algorithm}
\usepackage{xcolor}
\usepackage{enumitem}

\usepackage{booktabs}
\usepackage{multirow}
\usepackage{bm}
\usepackage{bbm}
\usepackage{amsmath}
\usepackage{amsmath}

\usepackage{algorithmic}
\usepackage{listings}
\usepackage{subfigure}
\usepackage{graphicx}
\usepackage{blindtext}
\usepackage{algorithm}

\usepackage{url} 

\makeatletter
\def\@copyrightspace{\relax}
\makeatother

\begin{document}

\title{\our: Root Cause Analysis with Partially Observed Data}


\author{
\IEEEauthorblockN{Chang Gong}
\IEEEauthorblockA{\textit{Institute of Computing Technology, } \\
\textit{Chinese Academy of Sciences}\\
gongchang21z@ict.ac.cn}

\\

\IEEEauthorblockN{Wenbin Li}
\IEEEauthorblockA{\textit{Institute of Computing Technology, } \\
\textit{Chinese Academy of Sciences}\\
liwenbin20z@ict.ac.cn}

\\

\IEEEauthorblockN{Kaiyu Feng}
\IEEEauthorblockA{\textit{Beijing Institute of Technology} \\
fengky@bit.edu.cn}

\and

\IEEEauthorblockN{Di Yao*\thanks{*Corresponding authors.}}
\IEEEauthorblockA{\textit{Institute of Computing Technology, } \\
\textit{Chinese Academy of Sciences}\\
yaodi@ict.ac.cn}

\\

\IEEEauthorblockN{Lanting Fang}
\IEEEauthorblockA{\textit{Beijing Institute of Technology} \\
lantingf@outlook.com}

\\

\IEEEauthorblockN{\\Peng Han}
\IEEEauthorblockA{\textit{University of Electronic Science and} \\
\textit{Technology of China}\\
penghan\_study@foxmail.com}

\and

\IEEEauthorblockN{Jin Wang}
\IEEEauthorblockA{\textit{Megagon Labs} \\
jin@megagon.ai}

\\

\IEEEauthorblockN{\\Yongtao Xie}
\IEEEauthorblockA{\textit{Southeast University} \\
220215512@seu.edu.cn}

\\

\IEEEauthorblockN{\\Jingping Bi*}
\IEEEauthorblockA{\textit{Institute of Computing Technology, } \\
\textit{Chinese Academy of Sciences}\\
bjp@ict.ac.cn}
}




\renewcommand{\algorithmicrequire}{ \textbf{Input:}} 
\renewcommand{\algorithmicensure}{ \textbf{Output:}} 

\newcommand{\ourTask}{{ourTask}\xspace}   

\newcommand{\our}{\texttt{PORCA}\xspace}  

\newcommand{\dfcdPart}{\texttt{Deconfounded Causal Learner}\xspace} 
\newcommand{\hacclPart}{\texttt{Heterogeneity-Aware Scheduler}\xspace} 
\newcommand{\trackPart}{\texttt{Spurious-Mitigated Root Tracker}\xspace}  

\newcommand{\ourConf}{\textsc{DKL-conf}\xspace} 
\newcommand{\ourRw}{\textsc{DKL-rw}\xspace} 
\newcommand{\ourSp}{\textsc{DKL-sp}\xspace} 
\newcommand{\ourUb}{\textsc{DKL-ub}\xspace} 



\newcommand{\ie}{\emph{i.e.}\xspace} 
\newcommand{\etal}{\emph{et~al.}\xspace} 
\newcommand{\etc}{\emph{etc.}\xspace} 
\newcommand{\eg}{\emph{e.g.}\xspace} 
\newcommand{\define}[3]{\vspace{1ex}\noindent{ \textbf{\textsc{Definition {#1}}} (#2): \emph{#3}\vspace{1ex}}
}
\newcommand{\hypothe}[3]{\vspace{1ex}\noindent{ \textbf{\textsc{Hypothesis {#1}}} (#2): \emph{#3}\vspace{1ex}}
}

\newcommand{\claim}[3]{\vspace{1ex}\noindent{ \textbf{\textsc{Lemma {#1}}} (#2): \emph{#3}\vspace{1ex}}
}

\newcommand\tab[1][0.5cm]{\hspace*{#1}}

\maketitle

\begin{abstract}

Root Cause Analysis (RCA) aims at identifying the underlying causes of system faults by uncovering and analyzing the causal structure from complex systems.
It has been widely used in many application domains.
Reliable diagnostic conclusions are of great importance in mitigating system failures and financial losses. 
However, previous studies implicitly assume a full observation of the system, which neglect the effect of partial observation (\ie, missing nodes and latent malfunction).
As a result, they fail in deriving reliable RCA results.

In this paper, we unveil the issues of unobserved confounders and heterogeneity in partial observation and come up with a new problem of root cause analysis with partially observed data. 
To achieve this, we propose \our, a novel RCA framework which can explore reliable root causes under both unobserved confounders and unobserved heterogeneity.
\our leverages magnified score-based causal discovery to efficiently optimize acyclic directed mixed graph under unobserved confounders. 
In addition, we also develop a heterogeneity-aware scheduling strategy to provide adaptive sample weights.
Extensive experimental results on one synthetic and two real-world datasets demonstrate the effectiveness and superiority of the proposed framework.
\end{abstract}


\section{Introduction}

Root Cause Analysis (RCA) aims to identify the underlining system failure so as to ensure the availability and reliability.
It has been widely used in many fields such as telecommunication~\cite{RCA_bg_telecom/zhang2020influence}, IT operations~\cite{RCA_Survey_2023_CSUR, RCA_Others_KDD22_CIRCA}, and manufacturing~\cite{RCA_bg_manufacturing_e2023automatic, DBLP:conf/icdm/HeTXWLLWTCK23}.
There are complex dependencies and interactions between components in systems of these fields.
Due to the large scale and complexity of such systems, they are vulnerable to failures, which could potentially lead to huge economic loss and degradation of user experiences. 
Many companies (\eg, Google, Amazon, Alibaba) have made huge efforts to apply RCA in system monitoring to address such issues.

\begin{figure}
    \centering
	\includegraphics[width=0.48\textwidth]{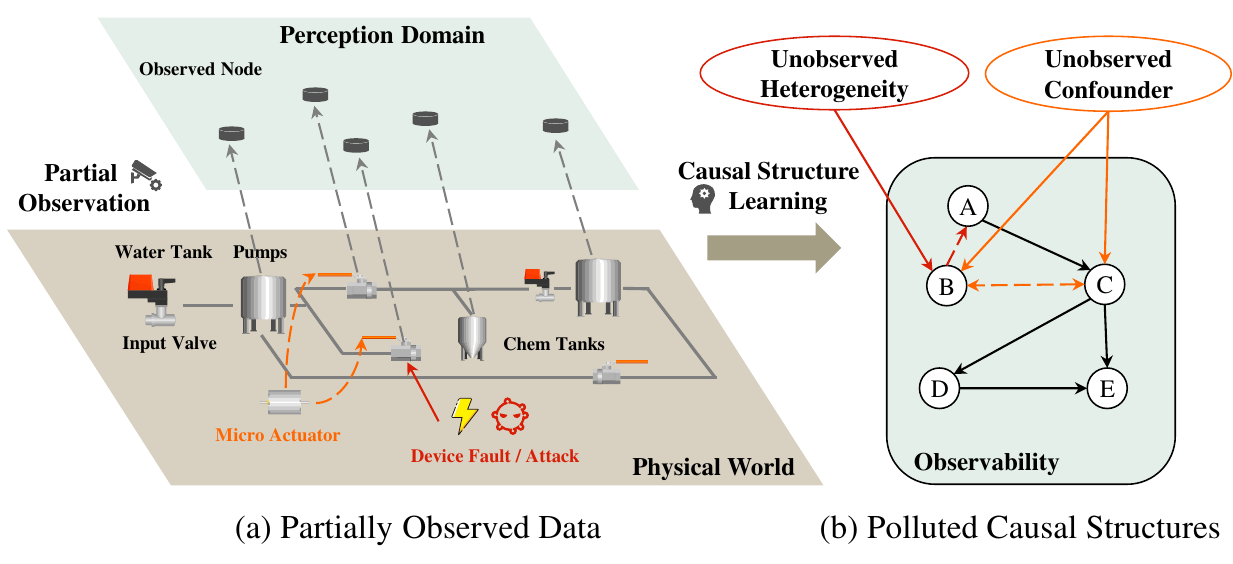}
	\vspace{-2ex} 
 	\caption{The motivation of \our. (a) shows that the physical world is partially observed, for example, micro actuators and latent malfunctions are neglected. (b) illustrates spurious edges caused by {unobserved confounders} and {{unobserved heterogeneity}} in {\color{orange}orange} and {\color{red}red}, respectively. }
	\vspace{-4ex}
        \label{fig:motivation}
\end{figure}

There is a long stream of research in both data mining~\cite{RCA_Others_KDD22_CIRCA, RCA_Others_KDD23_HRLFH, RCA_Baselines_KDD23_CORAL,  RCA_Others_ICDM17_causal_propagation} and software engineering~\cite{RCA_Baselines_IWQoS20_MicroCause} communities to ensure reliable RCA.
The objective is to provide reliable and robust localization for engineers to help address failures in a timely manner and avoid meaningless labours due to noisy false alarms. 
With the development of causal inference and trustworthy machine learning~\cite{causal_bg_causalSurvey/csur/GuoCLH020, DBLP:conf/icdm/KelenPKB23, gong2023causal, assaad2022survey} techniques, recent efforts build a causal dependency graph to represent system architecture and utilize causal discovery methods to boost the RCA pipeline~\cite{RCA_Baselines_KDD23_CORAL,RCA_Baselines_IWQoS20_MicroCause, RCA_Baselines_CCGRID18_CloudRanger, RCA_Baselines_NIPS22_RCD,  RCA_Others_KDD23_REASON}.
In such a graph, the nodes represent the performance metrics or services while edges represent the casual effects between them.
However, most of existing RCA approaches impose stringent assumption on causal sufficiency and neglect the existence and potential hazards of missing data.
The example in Figure~\ref{fig:motivation} shows a simplified illustration of a manufacturing testbed, the actuator can manipulate both two downstream pumps.
In many cases, small entities related human activities like this are often neglected or not monitored in system analysis~\cite{RCA_bg_leitmann1986feedback, RCA_bg_liu2016modeling}. 
Besides, the attack or latent malfunction on the pump node in Figure~\ref{fig:motivation} would change the data distribution and cause spurious edges in causal analysis. Similar issues also exist in a wide range of AIOps. 
For example, in microservice systems, biased models and misleading causal conclusion will result in false alarms and missing calls in root cause analysis~\cite{RCA_bg_CARE/eurosys/XuZLQPDLDZ21, RCA_Others_WWW23_CausIL}.
Some previous studies emphasize on the concept of ``observability" and leverage data engineering techniques, \eg, increasing the number of monitoring metrics or dive into metric, trace, and log data iteratively, to enhance perception domain~\cite{RCA_Survey_2023_CSUR}.
However, it is difficult to avoid such issues solely on the data level in real world application.

Witnessing the limitation of existing solutions, we argue that RCA should be able to automatically recognize and mitigate issues caused by partial observation on the model level.
To reach this goal, we need to address technical challenges from two aspects:\\

\noindent\textbf{Unobserved Confounders}.
In the partially observed data, the neglected entities in the system may serve as unobserved confounders and lead to spurious correlation in root cause analysis.
Without awareness of unobserved confounders, false edges and misleading causal conclusions may be derived, which will lead to false alarms in root cause analysis.
But the practical constraints and statistical limitations make it difficult to mitigate their effects.
On the one hand, vanilla structural causal models are misspecified in practice and might fail to account for latent nodes~\cite{CD_nohidden_NIPS18_NOTEARS_origin}.  
On the other hand, although some studies explore the additional statistical rules~\cite{causal_bg_10rules_zhang2008completeness} for uncovering relations with confounders, the heavy computational cost restricts their utilization in RCA.
\smallskip

\noindent\textbf{Unobserved Heterogeneity}.
Previous studies~\cite{CD_Rescore_zhang2022boosting, CD_DARING_hetroEVIDENCE_he2021daring} have proved that the heterogeneity of exogenous factors or noise distributions will lead to spurious edges and causal conclusions.
Nevertheless, it's nontrivial to account for heterogeneity in RCA, as they are latent to analyzers.
Although some previous studies try to alleviate this via manual labeling or rule-based change point detection~\cite{RCA_Baselines_KDD23_CORAL, RCA_Baselines_IWQoS20_MicroCause}, they still suffer from resource limitation or inflexiblility in dealing with different kinds of malfunctions and sophisticated attacks.
What's more, it is further challenging to discover the true causal structures under the circumstance of both unobserved confounders and unobserved heterogeneity~\cite{CD_hidden_AS_Dantzig_rothenhausler2019causal}.

To address these challenges, we come up with a new problem of Partially Observed Root Cause Analysis (\underline{\our}) that aims at tackling both issues and propose a comprehensive solution to it.
To deal with unobserved confounders, we first replace existing causal models that have strong assumptions with magnified structural causal models to avoid potential model misspecification.
We propose a magnified score function which allows for efficient causal discovery in a gradient-descent way.
In the aspect of unobserved heterogeneity, we distinguish normal and abnormal observations in a flexible manner via the distortion of causal mechanisms.
To be specific, we propose a heterogeneity-aware scheduling process to boost causal discovery, which distinguishes different observations and regulates the optimization procedures with adaptive weights.
Finally, we locate potential root causes by accounting for both node-level anomaly and anomaly propagation in deconfounded causal structures.
In a nutshell, the contributions of this paper are summarized as follows:
\begin{itemize}[leftmargin=4mm]
    \item We decompose the partial observation issues into unobserved confounders and unobserved heterogeneity, and define the new problem, \ie, Root Cause Analysis with Partially Observed data (\our).
    \item We propose a novel framework to systematically support the new task~\footnote{We will abuse the notion of \our to refer to the proposed framework if there is no ambiguity in the rest of this paper.}, with the technical contributions of magnified score-based causal discovery, heterogeneity-aware scheduling, and deconfounded root cause localization. And we also provide a theoretical guarantee of it.
    \item We conduct extensive experiments on one synthetic dataset and two datasets from real-world testbeds. Experimental results demonstrate the supiriority of the proposed framework.
\end{itemize}

\section{Related Work}

\subsection{Root Cause Analysis}

Root cause analysis aims to identify the underlying causes of system problem.
It shows great practical values and has been a popular research topic in a wide range of domains~\cite{ RCA_bg_telecom/zhang2020influence, RCA_Survey_2023_CSUR, RCA_bg_manufacturing_e2023automatic}.
Many causal approaches are introduced in root cause analysis to derive more reliable diagnostic conclusions~\cite{RCA_Others_KDD22_CIRCA, RCA_Baselines_IWQoS20_MicroCause}. 
Most of them follow a two stage framework, which first conduct causal discovery reflecting the complex dependencies within the system and then leverage downstream analysis to find out potential root causes~\cite{RCA_Baselines_IWQoS20_MicroCause, RCA_Baselines_CCGRID18_CloudRanger, RCA_Baselines_WWW20_AutoMAP}.
And some recent researches explore advanced topics, such as learning from human feedback~\cite{RCA_Others_KDD23_HRLFH}, and propagation on network of networks~\cite{RCA_Others_KDD23_REASON}.
However, few works noticed the issues originated from partial observations, \ie, unobserved confounders and unobserved heterogeneity.
As for the first issue, several endeavors~\cite{RCA_Baselines_IWQoS20_MicroCause, RCA_others_Seive_thalheim2017sieve} have mentioned the existence of unobserved confounders in RCA, but not accounted for it technically. 
Recently, Xu et al. ~\cite{RCA_bg_CARE/eurosys/XuZLQPDLDZ21} emphasize the effect of unobserved confounders. 
However, they focus on the design of experimental trail and debiasing, which is orthogonal to our work. 
As for the second issue, several studies~\cite{RCA_Baselines_KDD23_CORAL, RCA_Baselines_IWQoS20_MicroCause} try to identify data heterogeneity via rule-based change point detections while neglect the latent malfunctions. 
Chakraborty et al.~\cite{RCA_Others_WWW23_CausIL} alleviate heterogeneity by introducing external nodes, which relies on domain knowledge and cannot extend to general cases. 
Our study proposes a framework which systematically account for both unobserved confounders and unobserved heterogeneity in root cause analysis.

\subsection{Causal Discovery}
Causal discovery aims at uncovering causal structures from observational data.
By providing insights towards complex systems, the inferred structures will be beneficial to down stream tasks such as root cause analysis.
Existing works in causal discovery can be roughly divided into two categories, \ie, constraint-based methods~\cite{spirtes2000causation, entner2010causal}, and score-based methods~\cite{CD_nohidden_NIPS18_NOTEARS_origin, CD_hidden_AISTATS21_ABIC, DBLP:conf/icdm/XiaZRGZ23, DBLP:conf/clear2/LiuHGKBG24}. 
Compared to constraint-based methods, score-based methods could be seamlessly integrated with machine learning techniques and have better flexiblity and scalablity. 
Especially, owing to the recent efforts~\cite{CD_nohidden_NIPS18_NOTEARS_origin} converting the combinatorial search to continuous optimization, causal discovery can be conducted in an efficient way.
However, these methods assume causal sufficiency, restricting that there is no unobserved confounder.
Other works tackle unobserved confounders by leveraging additional rules~\cite{causal_bg_10rules_zhang2008completeness,  entner2010causal}, which encounter heavy computational overhead.
Some recent theoretical studies~\cite{CD_hidden_AISTATS21_ABIC} convert statistical property with the presence of unobserved confounders into numerical constraints.
To the best of our knowledge, no existing works in RCA could efficiently uncover causal structures with the presence of unobserved confounders.
Besides, the data in RCA is heterogeneous with latent attacks or malfunctions, which makes the issue more serious~\cite{CD_hidden_AS_Dantzig_rothenhausler2019causal}.

Heterogeneity in data would lead to spurious edges in causal discovery~\cite{CD_Rescore_zhang2022boosting, CD_DARING_hetroEVIDENCE_he2021daring}.
Most existing works for this issue assume the observability towards the heterogeneity in data. 
For example, Huang et al.~\cite{CD_CDNOD_huang2020causal} assume an external variable indicating the domain shift for heterogeneous data, and Lippe et al.~\cite{CD_intv_lippe2021efficient} assume the label information of intervention available, which cannot be easily applied to scenarios in RCA.
Unlike above previous studies, we address both two issues at the same time by proposing magnified score functions and scheduling causal discovery procedures with adaptive-weights.

\section{Preliminary}

\subsection{Causality Basics}

For the sake of clarity, we follow Pearl's framework~\cite{pearl2009causality} and first briefly describe basic concepts. 
Then, the idea of differentiable score-based causal discovery will be reviewed.

A causal graph represents the causal relations between variables as the directed acyclic graph (DAG) $\mathcal{G} = ( \mathcal{V}, \mathcal{E}  )$.
$\mathcal{V}$ denotes the set of nodes, where each node corresponds to a variable, for example, a system metric.
And $\mathcal{E} = \{ (\mathbf{v}_i, \mathbf{v}_j ) | \mathbf{v}_i, \mathbf{v}_j \in \mathcal{V} \} $ denotes the set of edges, where each edge $(\mathbf{v}_i, \mathbf{v}_j )$ represents a directional relationship from the causal variable $\mathbf{v}_i$ to the outcome variable $\mathbf{v}_j$.  
On this basis, Structural Causal Model (SCM) captures the asymmetry between causal direction on the data generation process.
To be specific, SCM with additive noise can be formalized as $\mathbf{x} = f (\mathbf{x}) + \textbf{u}$, where independent noise term $\textbf{u}$ reflects effect of the exogenous factor.     
Let metric $\mathbf{x}_i$ correspond to node $\mathbf{v}_i$, the mapping function $f(\cdot)$ could be deduced via the DAG $\mathcal{G}$. 
As a scalable framework with promising statistical properties~\cite{pearl2009causality}, SCM and its variants have been a mainstream methodology in causal machine learning~\cite{causal_bg_causalSurvey/tkdd/YaoCLLGZ21, causal_bg_causalSurvey/csur/GuoCLH020}.  

Causal discovery aims at uncovering the graph structure corresponding to SCM, which can provide valuable insights for further analysis.
Compared with previous studies, differentiable score-based causal discovery~\cite{CD_nohidden_NIPS18_NOTEARS_origin} can be seamlessly integrated with machine learning techniques and offers additional advantages, such as flexibility, scalability, and the ability to model nonlinear relations.
The basic idea is {(i)} to define a score function $\mathcal{S}$ that evaluates how well a given causal structure fits the observed data;
{(ii)} to specify the constraints $h(\cdot)$ such as acyclicity;
{(iii)} to derive causal conclusions by optimizing the score function, which can be conducted in a gradient-descent way. 

\subsection{Problem Definition}

To involve partially observed data into RCA, we need to explicitly account for unobserved confounders and heterogeneity. 
Previous studies use acyclic directed mixed graphs (ADMGs)~\cite{spirtes2000causation, CD_ADMG_Magnified_pena2016learning} with both directed and bidirected edges, where the latter indicates the existence of unobserved confounders.
Let $d$ be the number of observed metrics, $D \in \mathbbm{R}^{d\times d}$ and $B \in \mathbbm{R}^{d\times d}$ represent directed and bidirected adjacency matrix, respectively.  
In this paper, we turn the mixed graphs $(D,B)$ to magnified adjacency matrix form $M \in \mathbbm{R}^{(d+r)\times (d+r)}$, where $r$ is the number of latent nodes $\mathbf{C}$. To be specific, the $x_i \leftrightarrow x_j$ in $B$ can be represented as $c_k \rightarrow x_i$ and $c_k \rightarrow x_j$ in $M$.  
In addition, the noise term $\mathbf{u}$ has heterogeneous distribution with latent malfunction. 
To formally include above items, we leverage magnified SCM as Definition~1.

\define{1}{Magnified Structural Causal Model}{The partially observed data can be modulated as: 
$$[\mathbf{x}, \mathbf{c}] = f_M( \mathbf{x}, \mathbf{c}) + \mathbf{u},   $$  
where $f_M(\cdot)$ describes how observed and latent nodes being affected and can be transformed as magnified adjacency matrix $M$. And the noise term $\mathbf{u}$ follows heterogeneous distributions across different observations.} 

Based on the above definition, in this paper, the problem of \our can be formulated as following:
Given observed metrics $\mathbf{X} = (\textbf{x}_1^{1:T}, ..., \textbf{x}_d^{1:T})$, for alarm of front-end metric $y^t\ (\mathbf{y}\in\mathbf{X})$, we aim to build a model that (i) uncovers causal structures of the system by leveraging magnified SCM to simultaneously account for unobsreved confounders and heterogeneity;
(ii) identifies the top $K$ metrics in $\mathbf{X}$ as the potential root causes of $\mathbf{y}$.

\subsection{Method Overview}

\begin{figure*}
    \centering

	\includegraphics[width=0.85\textwidth]{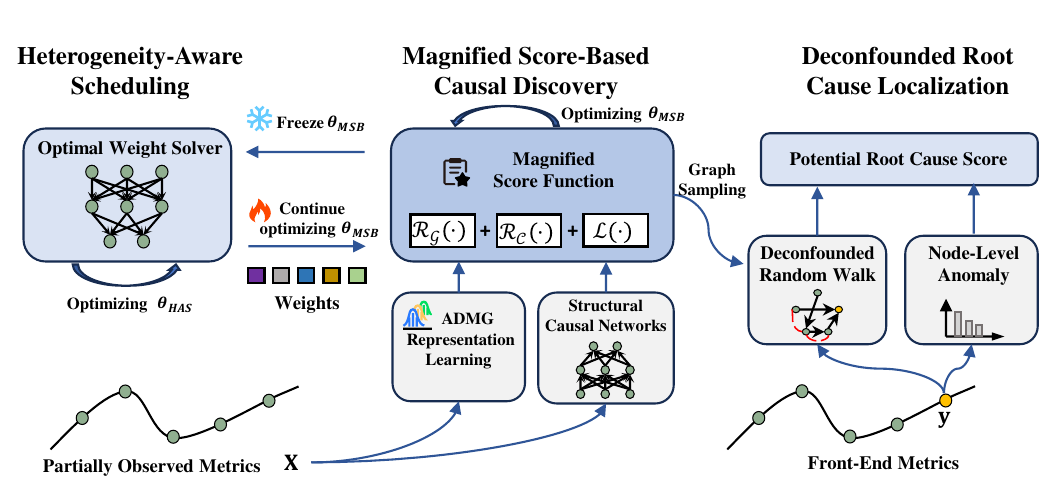} 
	\vspace{-4ex} 
	\caption{The overview of \our.}
	\vspace{-2ex}
         \label{fig:PORCA_overview}
\end{figure*}

As shown in Figure~\ref{fig:PORCA_overview}, the proposed \our framework contains three concise components: (i) magnified score-based causal discovery; (ii) heterogeneity-aware scheduling; and (iii) deconfounded root cause localization.
Assume $\theta_{\mathrm{MSB}}$ and $\theta_{\mathrm{HAS}}$ denote sets of learned parameters for magnified score-based causal discovery and heterogeneity-aware scheduling, respectively.
In the causal discovery component, we optimize $\theta_{\mathrm{MSB}}$ for magnified SCM.
In the scheduling component, given $\theta_{\mathrm{MSB}}$, we derive adaptive weights to schedule causal discovery procedures by optimizing $\theta_{\mathrm{HAS}}$.
In the localization component, potential root causes are ranked according to both node-level anomaly and anomaly propagation through deconfounded causal structures.

\section{Methodology}

\subsection{Magnified Score-Based Causal Discovery}

Given partially observed data $\mathbf{X} = (\textbf{x}_1^{1:T}, ..., \textbf{x}_d^{1:T})$,
magnified score-based causal discovery aims to continuously optimize parameters in maginified SCMs, which allows for inferring causal structures at the presence of unobserved confounders in a Bayesian perspective.
To this end, it first derives latent representations as posterior distribution for unobserved confounders $\mathbf{C}$, and the magnified graph $M$.
And it takes these representations and historical information as input to learn the system dynamics via neural networks.
Then the algebraic constraints~\cite{CD_nohidden_NIPS18_NOTEARS_origin,CD_hidden_AISTATS21_ABIC} on acyclic and ancestral properties is utilized for continuously optimization.    
This could be realized in three steps, \ie, 
(i) ADMG representation learning,
(ii) structural causal networks, 
and (iii) magnified score function.

\subsubsection{ADMG Representation Learning}
The magnified score-based causal discovery process is built on the basis of the ADMG structure $M$ and the unobserved confounders $\mathbf{C} = (\textbf{c}_1^{1:T}, ..., \textbf{c}_r^{1:T})$.
To parameterize them, we decompose the likelihood of an edge $p(M_{ij}  )$ via ENCO representation~\cite{CD_intv_lippe2021efficient}, \ie, ${\gamma}_{ij}$ and ${\theta}_{ij}$ denotes the existence and direction of edges, respectively.
Thus the posterior structural distribution to be approximated could be denoted as Equation~\eqref{eq:psd}:
\begin{equation}\label{eq:psd}
    q_{\gamma,\theta} ( M_{ij}  ) \sim \mathrm{Bern}( \mathrm{sigmoid}(\gamma_{ij}) \cdot \mathrm{sigmoid}(\theta_{ij}) ).
\end{equation}
As for the representation of unobserved confounders $\mathbf{C}$, we parameterize it as the mixture of Gaussian distributions in Equation~\eqref{eq:mgd}: 
\begin{equation}\label{eq:mgd}
    q(\mathbf{c}) \sim  \{ \mathcal{N}( \mu_i, \sigma_i^2) \}_{i=1}^r
\end{equation}
where the parameters $\mu_i$ and $\sigma_i^2$ are determined by MLP $f_{\mathrm{gauss}}(\mathbf{x})$.
The learned representation can be reckoned as $\mathbf{C}$'s posterior distribution given partial observations $\mathbf{X}$.

\subsubsection{Structural Causal Networks} 
We utilize structural causal networks to model the complex nonlinear mappings entailed in magnified SCM, \ie, $\mathbf{x}_j = f_{M,j}(\mathbf{x}, \mathbf{c} )$. 
According to the decoupling assumption~\cite{CD_hidden_AISTATS21_ABIC}, the formulation of structural causal network $f_{M, j}(\cdot)$ is detailed as Equation~\eqref{eq:SCN_1}:
\begin{equation}
        \mathbf{x}_j^t 
        = f_{ \mathrm{obs} ,j} \left(\sum_{i=1}^{d}  M_{ij}   {g}_{i} (\mathbf{x}_i^{<t})  \right)
        +f_{ \mathrm{conf} ,j}\left( \sum_{i=d+1}^{d+r}  M_{ij}  {g}_{i} (\mathbf{c}_{i-d}^{<t}) \right) . 
 \label{eq:SCN_1}   
\end{equation}
We follow~\cite{DBLP:conf/uai/AssaadDG22} and let $\mathbf{x}_i^{<t} = (..., x_i^{t-2}, x_i^{t-1}  )$ denote the temporal information of variable $i$.
And $f_{ \mathrm{obs} ,j}(\cdot)$ aggregates the observed variables' impact on target variable $j$, 
$f_{ \mathrm{conf} ,j}(\cdot)$ approximates the influence of unobserved confounders,
and $g_i(\cdot)$ introduces nonlinearity for the effect of each component $i$.
They can be implemented based on MLPs.

Let $\Omega$ denote parameters to be optimized in structural causal networks.
To reduce the search space of $\Omega$, we introduce shared weights for similar neural networks and trainable hidden representations indicating different variables.  
To be specific, suppose $\mathbf{z}$ denotes a trainbable hidden representation reflecting the target or input nodes,
we have $f_{ \mathrm{obs} ,j} (\cdot) = f_{ \mathrm{obs}} (\mathbf{z}_j,  \cdot),\  f_{ \mathrm{conf} ,j} (\cdot) = f_{ \mathrm{conf}} (\mathbf{z}_j,  \cdot), \ g_{i} (\cdot) = g (\mathbf{z}_i,  \cdot) $. 
Thus, the number of neural networks to be estimated in structural causal networks are reduced from $3d+r$ to $3$, where $d$ and $r$ are the numbers of observed and latent nodes, respectively. 

\subsubsection{Magnified Score Function}
The score function has three terms, \ie, the data-fitting term, the structural restriction, and the restriction for unobserved confounders. 
Thus, we have $\mathcal{S} = \mathcal{L} + \mathcal{R}_{\mathcal{G}} + \mathcal{R}_{\mathcal{C}} $.
The data-fitting term is given as Equation~\eqref{eq:data_fitting_term}:
\begin{equation}
    \mathcal{L}(\mathbf{x}, \mathbf{c}, M ) = \mathbbm{E}_{q_{\gamma,\theta}(M)}\left[\sum_{t=1}^{T}\left[\log p_{\Omega}(\mathbf{x}^{t}|\mathbf{x}^{<t}, \mathbf{c}^{<t}, M)\right]\right]. 
    \label{eq:data_fitting_term}
\end{equation}

The structural restriction term of $\mathcal{S}$ is implemented via KL divergence between the posterior distribution $p(M)$ and prior distribution $q_{\gamma, \theta}( M )$ over uncertain graph $M$ as Equation~\eqref{eq:sc}:
\begin{equation}\label{eq:sc}
    \mathcal{R}_{\mathcal{G}}(M) = -\mathrm{KL}\left[q_{\gamma, \theta}( M )||p( 
  M  )\right],
\end{equation}
Through prior distribution, sparsity penalty, acyclic constraints, and ancestral constraints can be added as Equation~\eqref{eq:all_constraint}:
\begin{equation}\label{eq:all_constraint}
    p(M)\propto\exp\left(-\lambda\| M\|_F^2-\rho h( M)^2-\alpha h( M)\right),
\end{equation}
where $\alpha$ and $\rho$ are increased while optimizing score functions in the augmented Lagrangian framework~\cite{Method_Related_CD_99_augLag}, $\lambda$ controls the sparse penalty.
And $h(M)$ is an extension of NOTEARS constraint~\cite{CD_nohidden_NIPS18_NOTEARS_origin}. 
We first decompose $M$ into $(D,B)$.
Following the theoretical results in the previous study~\cite{CD_hidden_AISTATS21_ABIC}, we set the algebraic constraint $h(M)$ as Equation~\eqref{eq:ac}:
\begin{equation}\label{eq:ac}
    h(M)=\mathrm{trace}\left(e^{D}\right)-d+\mathrm{sum}\left(e^{D} \circ {B}\right).
\end{equation}
Regarding the term of restriction for unobserved confounders $\mathcal{S}$, the KL divergence can be calculated as Equation~\eqref{eq:uc}:
\begin{equation}\label{eq:uc}
    \mathcal{R}_{\mathcal{C}}( \mathbf{x}, \mathbf{c}) = -\sum_{t=1}^T \mathrm{KL}    \left[q(\mathbf{c}^t|\mathbf{x}^t) ||p(\mathbf{c}^t)\right]  ,
\end{equation}
where the posterior distribution of $\mathbf{C}$ is derived based on ADMG representation learning.

\subsection{Heterogeneity-Aware Scheduling}

Next we introduce how to handle unobserved heterogeneity (\eg, latent attack, device fault), which could hurt the performance of causal discovery and the following process of root cause analysis.
The core idea is to develop a heterogeneity-aware scheduling approach.
It is inspired by curriculum learning which mocks the recognition of human-being to learn from easy tasks to hard ones~\cite{CL_survey_wang2021survey}.
It has been well explored in enhancing causal learning with heterogeneous data~\cite{CD_Rescore_zhang2022boosting, Method_Related_CL_nipsworkshop23_dahmani2023child}.

To be more specific, we reshape the optimization procedure of score-based causal discovery with adaptive sample weights.
Given learned parameters $\theta_{\mathrm{MSB}}$ for magnified SCM of an iteration, we first distinguish the heterogeneity of different samples via the goodness of data reconstruction. 
Then we concurrently derive optimal sample weights mitigating the influences of unobserved heterogeneity.

The observation at each time step can be distinguished via the data reconstruction term into easier and harder samples. 
Then we upweight the harder samples gradually to make the causal discovery procedures aware of more informative samples.
Before constructing the optimization framework with scheduling, we first assume the existence of oracle importance of each samples. Thus, we come up with the reweighted score functions defined in Equation~\eqref{eq:weight_M_score}:
\begin{equation}
    \mathcal{S}_w = \mathcal{L}_w + \mathcal{R}_{\mathcal{G}} + \mathcal{R}_{\mathcal{C},w}.
    \label{eq:weight_M_score}
\end{equation}

Here $\mathcal{L}_w$ and $\mathcal{R}_{\mathcal{C},w}$ are calculated via reweighted samples, as opposed to apply averaging scores on all samples equally:
\begin{equation}
    \begin{aligned}
    \mathcal{L}_w(\mathbf{x}, \mathbf{c}, M ) &= \mathbbm{E}_{q_{\gamma,\theta}(M)}\left[\sum_{t=1}^{T} w^t \left[\log p_{\Omega}(\mathbf{x}^{t}|\mathbf{x}^{<t}, \mathbf{c}^{<t}, M)\right]\right],  \nonumber \\ 
    \mathcal{R}_{\mathcal{C},w }( \mathbf{x}, \mathbf{c}) &= -\sum_{t=1}^T   w^t  \mathrm{KL}   \left[q(\mathbf{c}^t|\mathbf{x}^t) ||p(\mathbf{c}^t)\right] ,  \nonumber
\end{aligned}
\end{equation}
where $\textbf{w}=(w^1,...,w^T)$ and $w^t$ is the importance at time step $t$ denoting difficulty of heterogeneous observations.

In this process, we aim to embed the optimization of adaptive weights as scheduling to boost the original magnified score-based causal discovery procedure.
Formally, we have the bi-level optimization problem shown in Equation~\eqref{eq:bilevel}:
\begin{equation}
    {\max}_{ M, \mathbf{C} }  S_{\mathbf{w}^*}(\mathbf{x}, \mathbf{c}, M) \ \ \mathrm{s.t.~}\mathbf{w}^* \in\underset{\mathbf{w}\in\mathbbm{C}(\tau)}{\operatorname*{\arg\min}}S_{\mathbf{w}}(\mathbf{x}, \mathbf{c}, M).
    \label{eq:bilevel}
\end{equation}
where $\mathbbm{C}(\tau) = \{\mathbf{w}: 0  < \tau \leq w_1,...,w_T \leq \frac{1}{\tau}, \sum_{t=1}^T w_t = T  \} $,   
and the factor $\tau \in (0,1)$ represents the cutoff threshold for reweighting. 
The inner level objective in Equation~\eqref{eq:bilevel} is to minimize the reweighted score functions to optimize $\mathbf{w}$. 
To be specific, we leverage MLPs parameterized by $\theta_{HAS}$ to process the input $\mathbf{x}^t$ to produce corresponding sample weights $w$.
The outer level objective is to maximize the magnified score functions for estimating parameters in magnified SCM.
Solving the outer-level problem should be conditioned on the optimal value of the inner-level one.

\subsection{Deconfounded Root Cause Localization}
Given front-end metrics ${y}^t$ and learned magnified SCM, we aim to explore the potential root causes and return the top $K$ candidates.
Our deconfounded root cause localization algorithm achieves this goal by synergizing both propagation property in the topology structure and node-level anomaly scores in three steps: (i) deconfounded random walk, (ii) node-level anomalous rank, and (iii) potential root cause score.

\subsubsection{Deconfounded Random Walk}
The random walk algorithm is proved a good performance in capturing anomaly propagation~\cite{RCA_Baselines_KDD23_CORAL,  RCA_Baselines_IWQoS20_MicroCause}.
We adopt one of its variants to further mitigate the spurious correlation from unobserved confounders.
We first decompose $M$ into $(D,B)$, in which $D$ represents the causal dependency of anomaly propagation and $B$ entails spurious correlations due to unobserved confounders.
Then the learned matrix $D$ is transposed as $D^\top$.
We leverage random walk algorithm with restart from front-end metric on $D^\top$ to get probability score of each candidate node.
To be specific, the transition distribution $\mathbf{H}$ can be formulated as Equation~\eqref{eq:rw}:
\begin{equation}
    \mathbf{H}[i,j] = \frac  { (1-\phi) D^\top [i,j] } {\sum_{k=1}^d  D^\top [i,k]   }  ,
    \label{eq:rw}
\end{equation}
where $\phi \in [0,1]$ is the probability of jumping behaviour.
The walker stops after $N_{rw}$ steps, and each node is visited $\zeta_i$ times.   

\subsubsection{Node-level Anomalous Rank}
The causal Markov factorization in magnified SCM, \ie, $\mathbbm{P}(\mathbf{x}^t) = \prod_{i=1}^d \mathbbm{P}(x_i^t | \mathbbm{PA}(x_i^t)  ) $, allows for analyzing autonomous causal mechanism on the node-level~\cite{RCA_Others_ZhangKun22}.
We take the violation of causal mechanism as anomaly and rank the goodness of reconstruction of each mapping function $f_{M,i}(\mathbf{x}^{<t}, \mathbf{c}^{<t}   )$ as the anomaly degree $\eta_i^t$ of node $i$. Thus we have
\begin{equation}
    \eta_i^t = \mathrm{RANK}_i^t \left(\mathrm{log}\ p_{\Omega}(x_i^1 ) , ...,   \mathrm{log}\ p_{\Omega} (x_i^t ) , ..., \mathrm{log}\ p_{\Omega} (x_i^T )      \right).
    \label{eq:node_level}
\end{equation}
It compares $\mathrm{log}\ p_{\Omega}(x_i^t  )$ with all other observations from $[1,T]$ to quantify the violation of causal mechanism. 

\subsubsection{Potential Root Cause Score}
In this part, we need to consider both anomaly propagation $\zeta_i$ and node level anomaly degree $\eta_i^t$. 
Then we define the potential root cause score of node $i$ towards alarm at time $t$ as $s_i^t$ as Equation~\eqref{eq:potential_score}:
\begin{equation}
    s_i^t=\psi\bar{\zeta}_i+(1-\psi)\bar{\eta}_{i}^t,
    \label{eq:potential_score}
\end{equation}
where $\bar{\zeta}_i$ is the normalized ${\zeta}_i$ and $\bar{\eta}_{i}^t$ is the normalized ${\eta}_{i}^t$, respectively.
$\psi$ controls the contribution of propagation analysis and node-level anomaly degree.
The root causes at time $t$ are identified by picking up top $K$ scores $(s_1^t,...,s_i^t,...,s_d^t)$.

Finally we introduce the learning procedure of \our.
It starts with a warm-up stage. 
Then parameters for causal discovery are optimized, which is scheduled by adaptive weights.
At last, root cause localization is conducted to get final results.
The full process could be found in Algorithm~\ref{algri}.

\begin{algorithm}[t]
	\label{algri}
	\caption{Learning procedure of \our}
	
    \begin{algorithmic}[1]
     \REQUIRE ~~\\
	The observed metrics $\mathbf{X}$ and the alarm of front-end metric $y^t$ ; parameters $\theta_{\mathrm{MSB}} = \{ \Omega, \theta, \gamma\}$ and $\theta_{\mathrm{HAS}}$ \\
  
    \ENSURE ~~\\
     Top $K$ ranked nodes for potential root causes.    \\

    \STATE { {\textit{\# Causal Discovery and Scheduling} }} 
    
    \STATE \textbf{Initialize:} initialize scheduling module parameter $\theta_{HAS}$ to uniformly output $1$ for $\mathbf{w}$

    \FOR{ $l_1$ in RANGE(0, $L_{\mathrm{outer}}$)  }  
        \STATE Freeze scheduling module parameter $\theta_{\mathrm{HAS}}$
        \STATE Calculate $\mathbf{w}^*$ by applying the bound $[\tau, \frac{1}{\tau} ]$
        \STATE { {\textit{\# The outer-level optimization} }} 
        \STATE Optimize $\theta_{\mathrm{MSB}}$ by maximizing $S_{w^*}(\mathbf{x}, \mathbf{c}, M ) $
        \STATE { {\textit{\# Ignore scheduling in warm-up phase} }} 
        \IF{ $l_1 \geq  l_\mathrm{scheduling}$ }
            \FOR{$l_2$ in RANGE(0, $L_{\mathrm{inner}}  $) }
                \STATE Freeze causal discovery parameters $\theta_{\mathrm{MSB}}$
                \STATE Derive $\mathbf{w}$ via networks parameterized by $\theta_{\mathrm{HAS}}$ and apply the bound $ [\tau, \frac{1}{\tau} ]$
                \STATE { {\textit{\# The inner-level optimization} }} 
                \STATE Optimize $\theta_{\mathrm{HAS}}$ by minimizing $S_{w}(\mathbf{x}, \mathbf{c}, M ) $
            \ENDFOR
        \ENDIF    
    \ENDFOR

    \STATE { {\textit{\# Root Cause Localization} }} 
    \STATE Conduct random walk according to Eq.(\ref{eq:rw}) and get $\boldsymbol{\zeta}$
    \FOR{i in RANGE($0,d$)}
        \STATE Compute anomaly degree $\eta_i^t$ according to Eq.(\ref{eq:node_level})
        \STATE Compute potential scores $s_i^t$ according to Eq.(\ref{eq:potential_score})
    \ENDFOR
    

    \RETURN Top $K$ values of $\mathbf{s}^t$
    \end{algorithmic}
    \label{algri}
    \vspace{-1ex}
\end{algorithm}

\subsection{Theoretical Analysis of \our}

This section first briefly introduce Lemma 1 and its proof, which indicates identifiable ADMG can be recovered via optimizing magnified score function.
Following that, Lemma 2 is brought forth to prove the optimal characteristics of the weights employed in the scheduling process.
Lastly, the computational complexity of the proposed method is analyzed.

\claim{1}{ADMG's Identifiability with Magnified Score Function}{By assuming the effect of observed and unobserved nodes in the additive noise SCMs to be decoupled, i.e., $[\mathbf{x}, \mathbf{c}] = f_{D,\mathbf{x}}(\mathbf{x}, \Omega) + f_{B,\mathbf{c}}(\mathbf{c}, \Omega) +\mathbf{u}$, the ground-truth ADMG $M$ can be recovered when magnified score function $\mathcal{S}$ is optimized.}   

The proof of Lemma 1 consists of two decomposable parts, which we will briefly present.
(i) \textit{Structural identifiability of ADMG ($M$)}. Under the decoupled assumption, the previous results in~\cite{CD_hidden_AISTATS21_ABIC} can be extended that any potential situations of $M_{ij}=(D_{ij}, B_{ij})$, \ie, $(0,1), (1,0), (0,0)$ are distinguishable from the perspective of $p_{\Omega}(\mathbf{x}, M)$.  
(ii) \textit{Vanilla score function ($\mathcal{S}$) allows for efficiently recovering ADMG}. The score function $S$ serves as the evidence lower bound (ELBO) to be maximized,  \ie, $\mathcal{S} = \mathcal{L}_{\mathrm{ELBO}} \leq \sum_t \mathrm{log} p_{\Omega}(\mathbf{x}^t) $. To be specific, with the increases of observations, the graph restriction term $\mathcal{R}_{\mathcal{G}}$ asymptotically equals zero.
Let $p(\mathbf{x}, M^0)$ denote the data generation distribution with ground-truth ADMG $M^0$, the asymptotic estimation of the ELBO can be derived in Equation~\eqref{eq_proof_1_1}:
\begin{equation}
    \mathcal{L}_{\mathrm{ELBO}} =\int p(\mathbf{x}, M^0)
    \sum_{M \in\mathcal{M}_{\theta,\gamma}   }w_{\theta,\gamma}(M)\log p_\Omega(\mathbf{x}|M)d\mathbf{x}.
    \label{eq_proof_1_1}
\end{equation}
And the optimal value is achieved when the MLE solution $(\Omega^*, M^*)$ satisfies the Equation~\eqref{eq_proof_1_2}:
\begin{equation}
\mathbbm{E}_{p(\mathbf{x},M^0)}\left[\log p_{\Omega^*}(\mathbf{x}|M^*)\right]=\mathbbm{E}_{p(\mathbf{x},M^0)}\left[\log p(\mathbf{x}, M^0)\right].
    \label{eq_proof_1_2}
\end{equation}



\claim{2}{Optimal Property of Heterogeneity-Aware Weights}{In the scheduling phase, suppose that the observation at $t_1$ has a relative smaller data likelihood than that of $t_2$, i.e., $\mathcal{L}(\mathbf{x}^{t_1}  ) < \mathcal{L}(\mathbf{x}^{t_2}  )  $, where $t_1,t_2 \in \{1,...,T\}$. Then the heterogeneity-aware weights $w^{*,t_1}$ and $w^{*,t_2}$ have $w^{*,t_1} \geq w^{*,t_2}$. The equality only holds in thresholding conditions, i.e., $w^{*,t_1} = w^{*,t_2} = \tau$ or $w^{*,t_1} = w^{*,t_2} = 
\frac{1}{\tau}$.}

To prove Lemma 2, we additionally introduce a small permutation term $\epsilon$ on weights $w^*$, such that $\epsilon \in (0, \mathrm{min}(w^{*,t_1}-\tau, \frac{1}{\tau} -w^{*,t_2}   )  )  $. Then we replace $w^{*,t_1}$, $w^{*,t_2}$ in $w^*$ with permutated version $w'$, \ie, $(w^{*,t_1}+\epsilon)$ , $(w^{*,t_2})-\epsilon) \in w'$.
Thus, the contradiction of the reweighted score functions can be formed in Equation~\eqref{eq_proof_2}:
\begin{equation}
    \mathcal{S}_{w^*} - \mathcal{S}_{w'} = \epsilon \cdot [ \mathcal{L}(\mathbf{x}^{t_2}  ) - \mathcal{L}(\mathbf{x}^{t_1}  ) ] > 0,
    \label{eq_proof_2}
\end{equation}
which contradicts with $w^* \in \mathrm{arg\ min} \mathcal{S}_w(\mathbf{x},\mathbf{c},M) $.
Therefore, we have $w^{*,t_1} \geq w^{*,t_2}$ as stated in Lemma 2.

\noindent \textbf{Complexity analysis} \hspace{.5em}
We now analyze the computational complexity of each round in magnified score-based causal discovery.
Computing score function $\mathcal{S}$ and its gradient needs $O(WT)$ time, where $W$ is the number of parameters $\theta_{\mathrm{MSB}}$ to be optimized which scales according to a quadratic relationship with  $d$ and $r$.
For scheduling, the time complexity for getting optimal weights is $O(W'T)$, where $W'$ is the number of weights for the optimal weight solver.
As the running time of getting optimal $\theta_{\mathrm{MSB}}$ is larger than that of $\theta_{\mathrm{HAS}}$ to a great extent, the time complexity of the two processes is approximately $O(L_{outer}WT)$, where $L_{outer}$ is the iteration numbers for causal discovery with different optimal weights and $L_{outer} \leq 5$ in practice.
Taking the computation cost of root cause localization into account, let $N$ denote the number of iterations for random walk, the time complexity of the whole model is $O(L_{outer}WT + Nd^2  )$.







\section{Experiment}
In this section, we evaluate the performance of \our and answer the following questions: 
\begin{itemize}[leftmargin=4mm]
    \item \textbf{Q1:} What is the performance of \our in RCA compared with previous works without considering partially observed data?
    \item \textbf{Q2:} How could \our recover causal structures as intermediate results of RCA?
    \item \textbf{Q3:} What are the capabilities of the deconfounding module and heterogeneity-aware scheduling module?
    \item \textbf{Q4:} Is \our sensitive to parameter changes? 
\end{itemize}
Moreover, we provide a case study to illustrate the causal patterns entailed in \our can be beneficial to RCA in practice.

\subsection{Experimental Settings}

\renewcommand{\arraystretch}{1.6}
\begin{table*}[ht!] 
    \centering  
    \fontsize{9}{7.5}\selectfont
    \setlength\tabcolsep{3pt}
    \caption{RCA performance of all approaches in this study (bold: best; underline: runner-up).}
    \label{tb:RCA_results} \vspace{-1em}	
    \begin{tabular}{l|c|c|c|c|c|c|c|c|c}  
    \bottomrule
     & \multicolumn{3}{c|}{Simulation}  &  \multicolumn{3}{c|}{CRACs} & \multicolumn{3}{c}{SWaT} \cr
    \cline{2-10}
    &\ \ \ \ PR@5\ \ \ \ &\ \ PR@Avg\ \ & RankScore &\ \ \ \ PR@5\ \ \ \ &\ \ PR@Avg\ \ & RankScore &\ \ \ \ PR@5\ \ \ \ &\ \ PR@Avg\ \ & RankScore \cr
     \hline
    CloudRanger & 0.6750 & 0.5200 & 0.7649 & 0.6333 & 0.6299 & 0.7233 & 0.4305 & 0.2944 & 0.5384 \cr
    MicroCause & 0.8012 & 0.7356 & 0.8204 & 0.7017 & 0.6528 & 0.8033 & 0.5185 & 0.2916 & 0.5893 \cr
    AutoMAP & 0.6750 & 0.5399 & 0.7869 & 0.7075 & 0.6382 & 0.8267 & 0.4537 & 0.2740 & 0.5935 \cr
    CORAL & 0.8189 & 0.7125 & 0.8219 & 0.7383 & 0.6833 & 0.7947 & {0.6250} & \underline{0.4307} & \underline{0.6995} \cr
    RCD & 0.7967 & 0.7050 & 0.8437 & 0.7410 & 0.6667 & 0.8677 & 0.5416 & 0.3111 & 0.6287 \cr
    RCD* & \underline{0.8520}  & \underline{0.7783}  & 
  \underline{0.8633} & \underline{0.8040} & 0.7047 & 0.8832 & \underline{0.6311} & 0.3694 & 0.6583 \cr
    FCI* & 0.8359 & 0.7529 & 0.8602 & 0.7666 & \underline{0.7299} & \underline{0.9035} & 0.5879 & 0.4101 & 0.6129 \cr
    \hline
    \our & \textbf{0.9067} & \textbf{0.8199} & \textbf{0.9333} & \textbf{0.8333} & \textbf{0.7533} & \textbf{0.9267} & \textbf{0.6712} & \textbf{0.4324} & \textbf{0.7337} \cr
    \hline
    Improvement & 6.0\% & 5.1\% & 7.5\% & 3.5\% & 3.1\% & 2.5\% & 5.9\% & 0.3\% & 4.7\% \cr
    \toprule  
    \end{tabular}
    \vspace{-3ex}
\end{table*}

\subsubsection{Datasets}
We evaluate the performance of \our on both synthetic and real-world datasets. 

To remedy the problem that real-world datasets have no ground-truth of causal structure or explicitly labeled fault injection, we construct a synthetic dataset named \textbf{Simulation} that consists of $20$ nodes with $1000$ timesteps.
The causal structure is predefined, and the faults are injected by changing the edge weights and noise distributions. 
The details of the data generation procedure, including graph sampling, metrics generation, and fault injection are available in our code link.
With the propagation of causal networks, we can obtain the synthetic dataset. 
It allows us to explicitly and quantitatively account for both unobserved confounders and unobserved heterogeneity in the experimental evaluation. 

For the real-world scenarios, we choose two public datasets: 
\textbf{CRACs} is the monitoring data of a cooling system in a data center~\cite{datasets_li2023causal}. It consists of $38$ variables from January $1^{st}$, 2023 to May $1^{st}$, 2023. 
The occurrences and roots of abnormal temperature need to be identified for stable maintenance. 
\noindent\textbf{SWaT} is monitoring data collected from a real-world water treatment testbed~\cite{datasets_mathur2016swat}.
It consists of $51$ metrics from December $22^{nd}$, 2015 to January $2^{nd}$, 2016.
Physical and cyber attacks took place in the last four days.

\subsubsection{Experimental Protocol}
We evaluate the performance of \our on two tasks \ie, root cause analysis and causal discovery. 
In RCA experiment, the main results are the ranking based results on three datasets. 
For each dataset, we randomly mask $4$ nodes as the unobserved confounders. 
Moreover, we change the number of masked nodes to test the robustness of \our. 
In the causal discovery experiment, we report the performance of \our and other baselines on the synthetic and CRACs datasets whose causal structures are available. 

\subsubsection{Evaluation Metrics} 
We apply a variety of metrics for evaluating \our's performance.
For evaluating the performance of root cause analysis, we use \textbf{Precision@K (PR@K)}, \textbf{Precision@Average (PR@Avg)}, and \textbf{RankScore}.
\textbf{PR@K} indicates the number of correct root causes among top-K predictions, which is defined as: PR@K = $\frac{1}{\mathbbm{|A|} } \sum_{a \in \mathbbm{|A|} } \frac{ \sum_{i=1}^K  R_a(i)  }{ \mathrm{min}( K, |V_a| )  } $.
For $K=1,2,..,5$, we derive \textbf{PR@Avg} by averaging the values of PR@K.
\textbf{RankScore}, ranged from $0$ to $1$, is leveraged to evaluate the ranking ability of RCA methods.  
To be specific, RankScore $= \frac{1}{\mathbbm{|A|} } \sum_{a \in \mathbbm{|A|} }    \left( \frac{1}{ |R_a| } \sum_{R_a(i) \in V_a} \mathrm{score}(R_a(i))  \right)  $, where $\mathrm{score}(R_a(i)) = 1 - \frac{ \mathrm{max}(0, i - |V_a| )  }{ |R_a |  } $ if $R_a(i) \in V_a$, otherwise, the score equals zero.
As for evaluation on causal discovery results, two evaluation metrics are presented: \textbf{Area Under the Curve (AUC)} and \textbf{Structural Hamming Distance (SHD)}.
For aligning partial ancestral graphs and temporal causal graphs in the evaluation of causal discovery, we follow the similar setting as~\cite{CD_KK/icml/LiuK23}, (i) for usual directed edges $\rightarrow$, we treat them as regular edges; (ii) for bidirected edges $\leftrightarrow$, we drop them before evaluation; (iii) for other types of edges with ambiguous, only check whether skeleton is correct.


\subsubsection{Baselines}
We compare \our with seven representative baselines which can be categorized into two groups. 
The first group comprises existing RCA methods, such as CloudRanger\cite{RCA_Baselines_CCGRID18_CloudRanger}, MicroCause~\cite{RCA_Baselines_IWQoS20_MicroCause}, AutoMAP~\cite{RCA_Baselines_WWW20_AutoMAP}, CORAL~\cite{RCA_Baselines_KDD23_CORAL} and RCD~\cite{RCA_Baselines_NIPS22_RCD}.   
These methods are specifically designed for RCA without considering the influence of unobserved confounders and partially observed data. 
The second group consists of causal methods with unobserved confounders, such as RCD* and FCI*. 
To the best of our knowledge, existing solutions for RCA are incapable of modeling the unobserved confounders.
Thus, we extend RCD to the confounding settings, denoted by RCD*, for a fair comparison.
As for FCI*, we select the causal discovery method FCI~\cite{spirtes2000causation} to build the causal graph and conduct random walk to obtain the RCA results.

\renewcommand{\arraystretch}{1.6}
\begin{table}[ht!] 
    \centering  
    \fontsize{9}{7.5}\selectfont
    \setlength\tabcolsep{3pt}
    \caption{Performance of causal discovery (bold: best; underline: runner-up).}\vspace{-1em}
    \label{tb:CD_results} 	
    \begin{tabular}{l|c|c|c|c}  
    \bottomrule
     & \multicolumn{2}{c|}{Simulation}  &  \multicolumn{2}{c}{CRACs} \cr
    \cline{2-5}
    & \ \ \  AUC\ \ \   &  \  SHD\     &\ \  \  AUC\ \  \  &\   \ SHD\ \     \cr
     \hline
    CloudRanger(PC) & 0.8870 & 8 & 0.7259 & 57  \cr
    MicroCause(PCMCI) & 0.9268 & 7 & 0.7703 & 51  \cr
    AutoMAP(BGC) & 0.9157 & 8 & 0.7611 & 47  \cr
    CORAL(CRL) & 0.9453 & \underline{5} & 0.8091  & 39 \cr
    FCI & \underline{0.9477} & \underline{5} & \underline{0.8268} & \underline{33}  \cr
    \hline
    \our & \textbf{0.9759} & \textbf{2} & \textbf{0.8898} & \textbf{20}  \cr
    \toprule  
    \end{tabular}
    \vspace{-3ex}
\end{table}

\renewcommand{\arraystretch}{1.6}
\begin{table}[ht!] 

    \centering  
    \fontsize{9}{7.5}\selectfont
    \setlength\tabcolsep{3pt}
    \caption{Ablation-study results.}\vspace{-1em}
    \label{tb:ablation_new} 	
    \begin{tabular}{c|c|ccc}  
    \bottomrule
    Dataset & Method &\ \ \ \ PR@5\ \ \ \ &\ \ PR@Avg\ \ & RankScore \cr
    \hline

    \multirow{3}{*}{CRACs} 
    & w/o DC & 0.7257 & 0.6774 & 0.8275 \\
    & w/o SH & 0.7880 & 0.7191 & 0.8825 \\
    & \our & \textbf{0.8333} & \textbf{0.7533} & \textbf{0.9267} \\
    \hline
    \multirow{3}{*}{SWaT} 
    & w/o DC & 0.6296 & 0.4067 & 0.6523 \\
    & w/o SH & 0.6620 & 0.4205 & 0.7081 \\
    & \our & \textbf{0.6712} & \textbf{0.4324} & \textbf{0.7337} \\
    \toprule  
    \end{tabular}
    \vspace{-3ex}
\end{table}

\subsection{Performance Comparison of RCA}
\label{subsec:RCA}

To answer {Q1}, we compare \our with baseline methods in RCA.
From the results shown in Table~\ref{tb:RCA_results}, we have the following observations:

Firstly, \our outperforms all the compared baselines by a large margin. For example, the improvements of PR@5 on the synthetic dataset are over $6$\% while the absolute performance is over $90$\%. 
It proves that the score-based causal discovery and heterogeneity-aware scheduling techniques are effective in modeling the unobserved confounders. 
Secondly, the methods accounting for the effect of unobserved nodes (\eg, RCD* and FCI*) outperform other approaches assuming causal sufficiency (\eg, CloudRanger, MicroCause, AutoMAP, CORAL, and RCD). 
We can see that RCD* is the ablation of RCD which only replaces the causal discovery method from $\phi$-PC with $\phi$-FCI. 
The PR@5 improvement on CRACs dataset is about $6$\% compared with its original version. 
This is because unobserved nodes may lead to spurious correlations raising the issue of false alarms in root cause analysis.
Thirdly, CORAL is the strongest baseline among methods with causal sufficiency. 
By explicitly leveraging the changes of distribution in causal modeling, CORAL can update the causal graph incrementally, which could capture the heterogeneity data generation mechanism. 
Due to the absence of unobserved confounder modeling, it is also inferior to RCD* and FCI*, \ie, the PR@5 of CORAL on CRACs dataset is $5$\% less than RCD*.
Lastly, although RCD* achieves the best performance of all the baselines, it is also inferior to \our because of its incomplete ability to capture data heterogeneity. 
The heterogeneity-aware scheduling module in \our explicitly takes into account the varying distribution caused by latent malfunction in causal structure learning. 
Such a process can mitigate the spurious edges effect from heterogeneous data and learn more reliable causal edges to enhance the root localization performance. 
Thus, the experimental results demonstrate the superiority and effectiveness of \our in root cause analysis over other baseline methods.

\subsection{Performance Comparison of Causal Discovery}
\label{subsec:CD}

As the answer to Q2, we compare the causal discovery modules of baselines menioned above with \our on the Simulation and CRACs datasets. Other settings are the same with our RCA experiments. 
The results are shown in Table~\ref{tb:CD_results} where the involved casual discovery methods are attached after the baselines. 
We can observe that: 
(i) \our performs best with the highest AUC (0.9759) and the lowest SHD (2) on the Simulation dataset, indicating that it is most accurate in both metrics.
FCI is the seconde best method with a slightly lower AUC (0.9477) and the same SHD (5) as CORAL. Similar results can be observed in the CRACs dataset.
(ii) Compared to methods with causal sufficiency assumptions, FCI* and \our achieve better performance compared with other baselines. For example, the AUC of the best competitor CORAL is also $8$\% less than \our on CRAC dataset. This phenomena could be attributed to their ability to handle unobserved confounders leading to more accurate causal models when such confounders are present in the data. 
(iii) Among all the compared methods, the causal representation learning method CORAL (denoted as CRL in the table) achieves the best performance. By learning representations that capture the underlying causal structure, the algorithm could better distinguish causal from non-causal relationships. 
The result of CORAL is still inferior to \our, indicating the effectiveness of the proposed techniques. 

\subsection{Ablation Study}
\label{subsec:ab_para}

As the answer to Q3, we compare \our with two ablation methods, \ie, w/o DC and w/o SH. 
The first ablation, w/o DC, involves removing the deconfounding component,
and the second ablation, w/o SH, involves removing the scheduling component.
As shown in Table~\ref{tb:ablation_new}, both methods perform worse than \our under all settings, which suggests that both ``DC" and ``SH" contribute positively to the algorithm's performance. 
The improvement is consistent and significant, indicating that the combination of deconfounding and scheduling modules in \our is essential to obtain the optimal performance. 
We also see that the performance of w/o SH is higher than w/o DC. 
For example, on CRACs dataset, w/o SH achieves about $6$\% improvement on PR@5 compared with w/o DC. 
This phenomena indicates that the ``DC" component seems particularly important, as the performance drop is more significant when ``DC" is removed. 

\subsection{Parameter Analysis} 
\begin{figure}[!t]
\centering

\subfigure[Precision@1 for Simulation]{
\includegraphics[width=4.1cm]{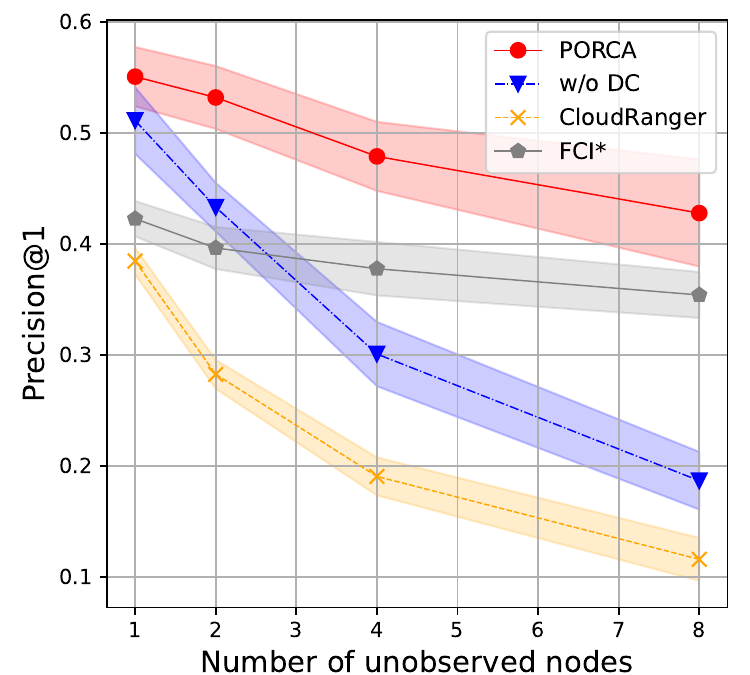}
}
\hspace{-3mm}
\subfigure[AUC for Simulation]{
\includegraphics[width=4.1cm]{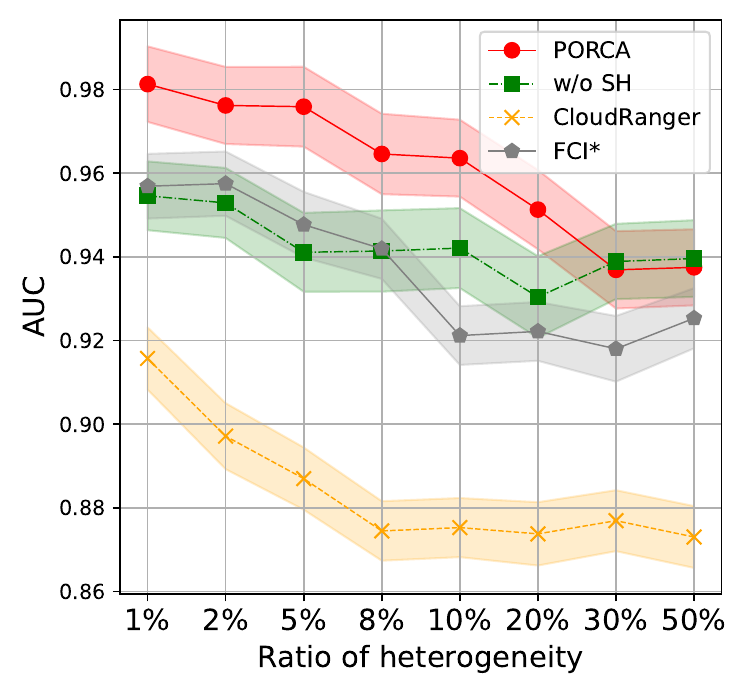}
}
\vspace{-9pt}

\caption{
Robustness of \our.}
\label{fig:ab_big}
\vspace{-2ex}
\end{figure}

To answer Q4, we conduct the parameter analysis.
To unveil the robustness of \our, we first utilize Simulation dataset, which can explicitly leverage the number of unobserved nodes and ratio of heterogeneity.
The results are illustrated in Figure~\ref{fig:ab_big}.
We can observe that:
(i) In Figure~\ref{fig:ab_big}(a), when the number of unobserved node varies among $\{1,2,4,8 \}$, \our's performance in terms of PR@1 is more robust than the ablation method without deconfounding module. 
To be specific, the proposed \our can get high first-shot prediction results within $4$ unobserved nodes, and endures a small fraction of drop as the unobserved nodes increase.
It demonstrates the robustness of \our with varying unobserved nodes.
(ii) When varying the ratio of heterogeneous observations, the performance comparison in terms of AUC between \our and the ablation version without scheduling module is shown in Figure~\ref{fig:ab_big}(b).
The reason that the compared method outperforms \our is that the heterogeneous data degenerates to balanced data with two 
latent categories.
Meanwhile, \our has a better performance when ratio is less than $30\%$, which is totally acceptable in real-world scenarios where latent malfunctions only occur a small fraction of time.

\begin{figure}[!t]
\centering

\subfigure[Model performance \textit{w.r.t.} $\lambda$]{
\includegraphics[width=4.35cm]{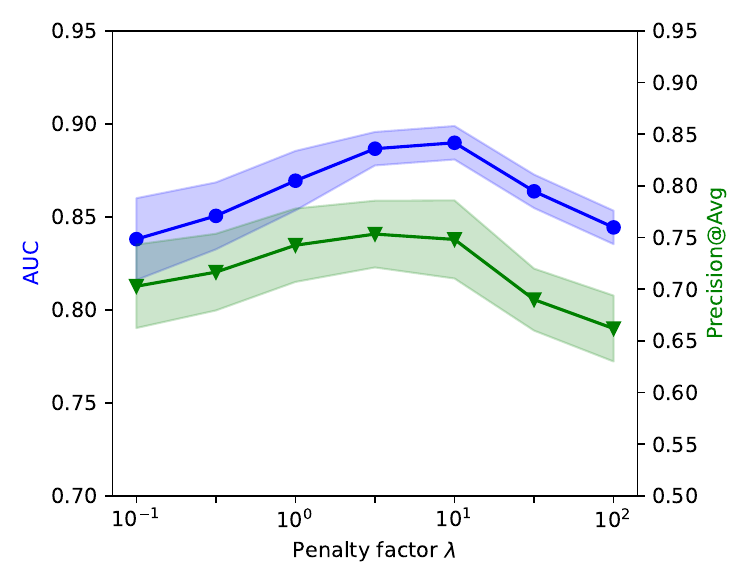}

}
\abovecaptionskip=4pt
\hspace{-6mm}
\subfigure[Model performance \textit{w.r.t.} $r$]{
\includegraphics[width=4.35cm]{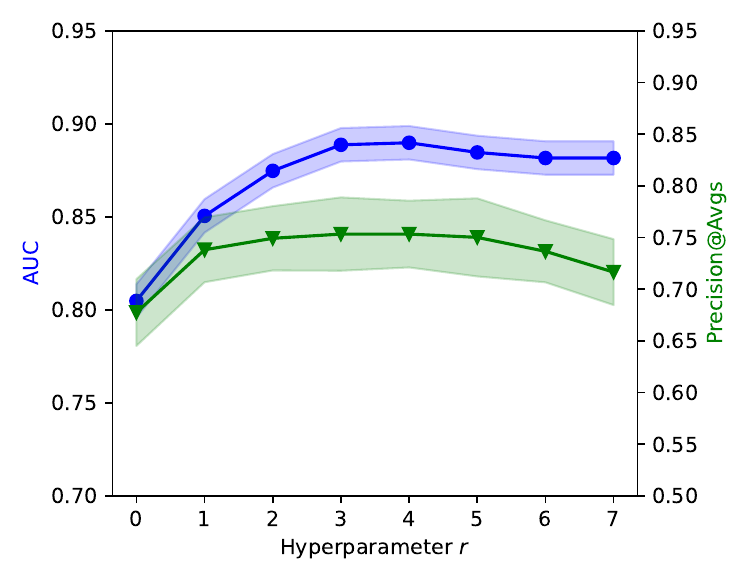}
}
\vspace{-11pt}

\caption{
Parameter analysis of \our.}
        \label{fig:para}
\vspace{-3ex}
\end{figure}

Then Figure~\ref{fig:para} illustrates the results of parameter analysis on CRACs dataset.
The factor $\lambda$ modifies the sparse penalty of the structure.
According to Figure~\ref{fig:para}(a), as $\lambda$ increases, the results of AUC and PR@Avg both increase steadily and peak around $5-10$. 
It shows that we can control the fitness of causal structures and reasonable values by adjusting $\lambda$ so as to benefit root cause localization.
Hyperparameters $r$ modifies the number of unobserved confounders $\mathbf{c}$ in magnified SCM.
In Figure~\ref{fig:para}(b), we witness that the performance of PORCA is robust even when $r$ is misspecified. 
This is due to that there isn't a strict one-to-one correlation between the matrix $M$ and observed spurious correlations. This non-isomorphism confers a degree of fault tolerance to hyperparameter $r$ misspecification, which is meaningful in real-world practice.

\subsection{Case Study}

\begin{figure}[!t]

\centering

\subfigure[Causal structure from \our]{
\includegraphics[width=4.15cm]{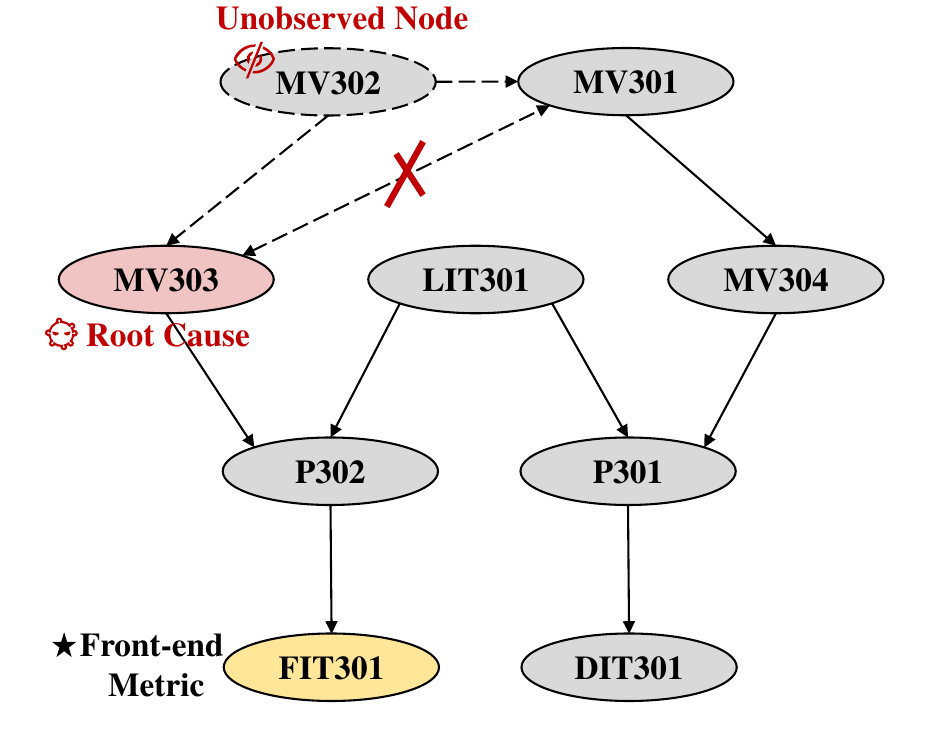}
}
\hspace{-3mm}
\subfigure[Causal structure from AutoMAP]{
\includegraphics[width=4.15cm]{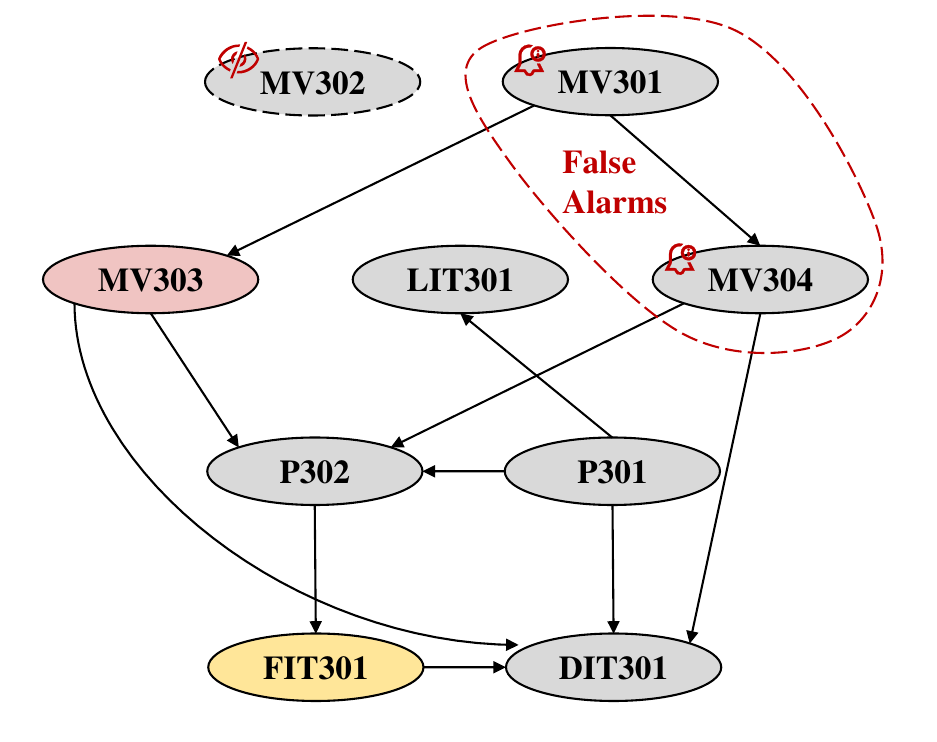}
}
\vspace{-9pt}

\caption{Learned causal structures from SWaT dataset. Subgraphs for stage 3 are illustrated. } 
\label{fig:case_deconf}
\vspace{-3ex}
\end{figure}

\begin{figure}
    \centering

	\includegraphics[width=0.45\textwidth]{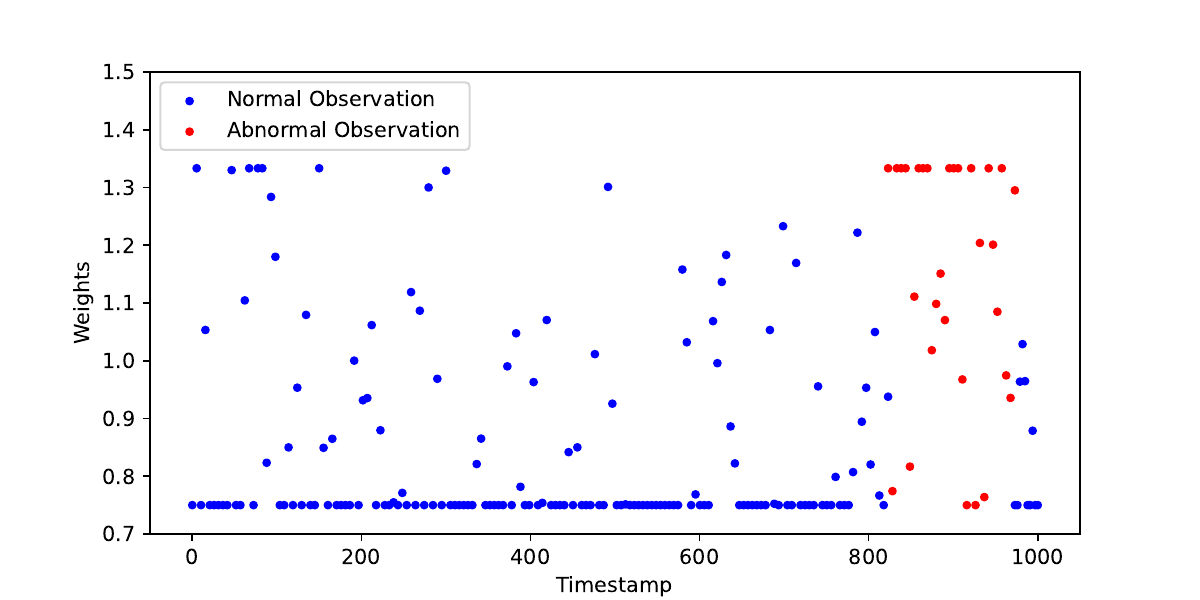}
 \vspace{-2ex}
	\caption{Illustration of learned weights by \our. Labels of abnormal observations are dotted in {\color{red}red} according to fault injection process. And the sampling frequency to $5$ for visualization.}
	\vspace{-3ex}
        \label{fig:case_weighting}
\end{figure}

Finally, we conduct two case studies to show the effectiveness of \our in practice.
Figure~\ref{fig:case_deconf} shows the learned causal structures from a stage in SWaT to explore the potential ability of \our in RCA with the presence of unobserved nodes. 
And Figure~\ref{fig:case_weighting} illustrates the derived weights from Simulation dataset with latent malfunction explicitly visible in drawing. 

We first conduct case studies on SWaT, and focus on the interaction at stage 3 (ultrafiltration) in the testbed.
In this case, the motorised valve \textit{MV303} is under single point attack, and the sensor \textit{FIT301} measuring the output to the next stage serve as the front-end metric. 
We mask the observation from \textit{MV302}, which is reported to have control dependencies on many downstream components~\cite{dataset_CD4swat_maiti2023iccps} (\eg, \textit{MV302}$\to$\textit{MV301} and \textit{MV302}$\to$\textit{MV303}), to explicitly reproduce the situation with unobserved confounders. 
A subgraph of the learned causal structure from \our is illustrated in Figure~\ref{fig:case_deconf}(a), while in Figure~\ref{fig:case_deconf}(b) the counterpart from AutoMAP is visualized.
We can observe that: 
(i) Insights about system interactions can be entailed from causal patterns. 
As illustrated above, two outputs of stage 3 depend on two pumps, which are determined by hierarchical control dependency from a series motorised valves. 
(ii) The presence of unobserved confounder (\textit{MV302}) does not lead to spurious edges. 
(iii) According to \our, \textit{MV303} is the one most likely to influence the front-end metric.
These observations show that \our can accurately locate the root cause via mitigating the effect of unobserved confounders.
Compared with \our, AutoMAP and other baselines fail to discover these interactions and may lead to false alarms due to spurious correlations.

Besides, Figure~\ref{fig:case_weighting} illustrates the learned weights by heterogeneity-aware scheduling across $1000$ steps on Simulation data.
We can observe that heterogeneous observations are distinguished with adaptive weight, which demonstrates the effectiveness of \our under heterogeneous circumstances.

\section{Conclusion}

In this paper, we define the problem of Root Cause Analysis with Partially Observed data (\our), which is essential in ensuring availability and reliability.
To tackle the issues of unobserved confounders and unobserved heterogeneity, we propose a brand new framework as the solution.
We prove that \our is capable of identifying the true causal structures with unobserved confounders, and the learned weights can correspond to the heterogeneity. 
We conduct an extensive set of experiments on the synthetic dataset and real-world data from two testbeds.
The results show that \our outperform all the compared baselines with partially observed data and works well in real-world applications.


\bibliographystyle{IEEEtran}
\bibliography{references}

\begin{thebibliography}{10}
\providecommand{\url}[1]{#1}
\csname url@samestyle\endcsname
\providecommand{\newblock}{\relax}
\providecommand{\bibinfo}[2]{#2}
\providecommand{\BIBentrySTDinterwordspacing}{\spaceskip=0pt\relax}
\providecommand{\BIBentryALTinterwordstretchfactor}{4}
\providecommand{\BIBentryALTinterwordspacing}{\spaceskip=\fontdimen2\font plus
\BIBentryALTinterwordstretchfactor\fontdimen3\font minus \fontdimen4\font\relax}
\providecommand{\BIBforeignlanguage}[2]{{%
\expandafter\ifx\csname l@#1\endcsname\relax
\typeout{** WARNING: IEEEtran.bst: No hyphenation pattern has been}%
\typeout{** loaded for the language `#1'. Using the pattern for}%
\typeout{** the default language instead.}%
\else
\language=\csname l@#1\endcsname
\fi
#2}}
\providecommand{\BIBdecl}{\relax}
\BIBdecl

\bibitem{RCA_bg_telecom/zhang2020influence}
K.~Zhang, M.~Kalander, M.~Zhou, X.~Zhang, and J.~Ye, ``An influence-based approach for root cause alarm discovery in telecom networks,'' in \emph{International Conference on Service-Oriented Computing}.\hskip 1em plus 0.5em minus 0.4em\relax Springer, 2020, pp. 124--136.

\bibitem{RCA_Survey_2023_CSUR}
J.~Soldani and A.~Brogi, ``Anomaly detection and failure root cause analysis in (micro) service-based cloud applications: {A} survey,'' \emph{{ACM} Comput. Surv.}, vol.~55, no.~3, pp. 59:1--59:39, 2023.

\bibitem{RCA_Others_KDD22_CIRCA}
M.~Li, Z.~Li, K.~Yin, X.~Nie, W.~Zhang, K.~Sui, and D.~Pei, ``Causal inference-based root cause analysis for online service systems with intervention recognition,'' in \emph{{KDD} '22: The 28th {ACM} {SIGKDD} Conference on Knowledge Discovery and Data Mining}, 2022, pp. 3230--3240.

\bibitem{RCA_bg_manufacturing_e2023automatic}
E.~e~Oliveira, V.~L. Migu{\'e}is, and J.~L. Borges, ``Automatic root cause analysis in manufacturing: an overview \& conceptualization,'' \emph{Journal of Intelligent Manufacturing}, vol.~34, no.~5, pp. 2061--2078, 2023.

\bibitem{DBLP:conf/icdm/HeTXWLLWTCK23}
C.~He, F.~Tian, P.~Xue, Y.~Wu, Y.~Li, J.~Li, Z.~Wang, F.~Tan, H.~Chen, and L.~Kong, ``Patternrca: {A} pattern-aware root cause analysis framework for multi-dimensional time series,'' in \emph{{IEEE} International Conference on Data Mining, {ICDM} 2023}, 2023, pp. 1583--1592.

\bibitem{RCA_Others_KDD23_HRLFH}
L.~Wang, C.~Zhang, R.~Ding, Y.~Xu, Q.~Chen, W.~Zou, Q.~Chen, M.~Zhang, X.~Gao, H.~Fan, S.~Rajmohan, Q.~Lin, and D.~Zhang, ``Root cause analysis for microservice systems via hierarchical reinforcement learning from human feedback,'' in \emph{Proceedings of the 29th {ACM} {SIGKDD} Conference on Knowledge Discovery and Data Mining, {KDD} 2023}, 2023, pp. 5116--5125.

\bibitem{RCA_Baselines_KDD23_CORAL}
D.~Wang, Z.~Chen, Y.~Fu, Y.~Liu, and H.~Chen, ``Incremental causal graph learning for online root cause analysis,'' in \emph{Proceedings of the 29th {ACM} {SIGKDD} Conference on Knowledge Discovery and Data Mining, {KDD} 2023}, 2023, pp. 2269--2278.

\bibitem{RCA_Others_ICDM17_causal_propagation}
J.~Ni, W.~Cheng, K.~Zhang, D.~Song, T.~Yan, H.~Chen, and X.~Zhang, ``Ranking causal anomalies by modeling local propagations on networked systems,'' in \emph{2017 {IEEE} International Conference on Data Mining, {ICDM} 2017}, 2017, pp. 1003--1008.

\bibitem{RCA_Baselines_IWQoS20_MicroCause}
Y.~Meng, S.~Zhang, Y.~Sun, R.~Zhang, Z.~Hu, Y.~Zhang, C.~Jia, Z.~Wang, and D.~Pei, ``Localizing failure root causes in a microservice through causality inference,'' in \emph{28th {IEEE/ACM} International Symposium on Quality of Service, IWQoS 2020}, 2020, pp. 1--10.

\bibitem{causal_bg_causalSurvey/csur/GuoCLH020}
R.~Guo, L.~Cheng, J.~Li, P.~R. Hahn, and H.~Liu, ``A survey of learning causality with data: Problems and methods,'' \emph{{ACM} Comput. Surv.}, vol.~53, no.~4, pp. 75:1--75:37, 2021.

\bibitem{DBLP:conf/icdm/KelenPKB23}
D.~M. Kelen, M.~Petreczky, P.~Kersch, and A.~A. Bencz{\'{u}}r, ``Theoretical evaluation of asymmetric shapley values for root-cause analysis,'' in \emph{{IEEE} International Conference on Data Mining, {ICDM} 2023}, 2023, pp. 210--219.

\bibitem{gong2023causal}
C.~Gong, D.~Yao, C.~Zhang, W.~Li, J.~Bi, L.~Du, and J.~Wang, ``Causal discovery from temporal data,'' in \emph{Proceedings of the 29th ACM SIGKDD Conference on Knowledge Discovery and Data Mining}, 2023, pp. 5803--5804.

\bibitem{assaad2022survey}
C.~K. Assaad, E.~Devijver, and E.~Gaussier, ``Survey and evaluation of causal discovery methods for time series,'' \emph{Journal of Artificial Intelligence Research}, vol.~73, pp. 767--819, 2022.

\bibitem{RCA_Baselines_CCGRID18_CloudRanger}
P.~Wang, J.~Xu, M.~Ma, W.~Lin, D.~Pan, Y.~Wang, and P.~Chen, ``Cloudranger: Root cause identification for cloud native systems,'' in \emph{18th {IEEE/ACM} International Symposium on Cluster, Cloud and Grid Computing, {CCGRID} 2018}, 2018, pp. 492--502.

\bibitem{RCA_Baselines_NIPS22_RCD}
A.~Ikram, S.~Chakraborty, S.~Mitra, S.~K. Saini, S.~Bagchi, and M.~Kocaoglu, ``Root cause analysis of failures in microservices through causal discovery,'' in \emph{Annual Conference on Neural Information Processing Systems 2022, NeurIPS 2022}, 2022.

\bibitem{RCA_Others_KDD23_REASON}
D.~Wang, Z.~Chen, J.~Ni, L.~Tong, Z.~Wang, Y.~Fu, and H.~Chen, ``Interdependent causal networks for root cause localization,'' in \emph{Proceedings of the 29th {ACM} {SIGKDD} Conference on Knowledge Discovery and Data Mining, {KDD} 2023}, 2023, pp. 5051--5060.

\bibitem{RCA_bg_leitmann1986feedback}
G.~Leitmann, E.~Ryan, and A.~Steinberg, ``Feedback control of uncertain systems: robustness with respect to neglected actuator and sensor dynamics,'' \emph{International Journal of Control}, vol.~43, no.~4, pp. 1243--1256, 1986.

\bibitem{RCA_bg_liu2016modeling}
J.~Liu, X.~Zhang, and X.~Chen, ``Modeling and active vibration control of a coupling system of structure and actuators,'' \emph{Journal of Vibration and Control}, vol.~22, no.~2, pp. 382--395, 2016.

\bibitem{RCA_bg_CARE/eurosys/XuZLQPDLDZ21}
Y.~Xu, X.~Zhang, C.~Luo, S.~Qin, R.~Pandey, C.~Du, Q.~Lin, Y.~Dang, and A.~Zhou, ``{CARE:} infusing causal aware thinking to root cause analysis in cloud system,'' in \emph{{HAOC} '21: Proceedings of the 1st Workshop on High Availability and Observability of Cloud Systems}, 2021, pp. 1--3.

\bibitem{RCA_Others_WWW23_CausIL}
S.~Chakraborty, S.~Garg, S.~Agarwal, A.~Chauhan, and S.~K. Saini, ``Causil: Causal graph for instance level microservice data,'' in \emph{Proceedings of the {ACM} Web Conference 2023, {WWW} 2023}, pp. 2905--2915.

\bibitem{CD_nohidden_NIPS18_NOTEARS_origin}
X.~Zheng, B.~Aragam, P.~Ravikumar, and E.~P. Xing, ``Dags with {NO} {TEARS:} continuous optimization for structure learning,'' in \emph{Advances in Neural Information Processing Systems 31: Annual Conference on Neural Information Processing Systems 2018, NeurIPS 2018, December 3-8, 2018, Montr{\'{e}}al, Canada}, 2018, pp. 9492--9503.

\bibitem{causal_bg_10rules_zhang2008completeness}
J.~Zhang, ``On the completeness of orientation rules for causal discovery in the presence of latent confounders and selection bias,'' \emph{Artificial Intelligence}, vol. 172, no. 16-17, pp. 1873--1896, 2008.

\bibitem{CD_Rescore_zhang2022boosting}
A.~Zhang, F.~Liu, W.~Ma, Z.~Cai, X.~Wang, and T.-S. Chua, ``Boosting causal discovery via adaptive sample reweighting,'' in \emph{The Eleventh International Conference on Learning Representations}, 2022.

\bibitem{CD_DARING_hetroEVIDENCE_he2021daring}
Y.~He, P.~Cui, Z.~Shen, R.~Xu, F.~Liu, and Y.~Jiang, ``Daring: Differentiable causal discovery with residual independence,'' in \emph{Proceedings of the 27th ACM SIGKDD Conference on Knowledge Discovery \& Data Mining}, 2021, pp. 596--605.

\bibitem{CD_hidden_AS_Dantzig_rothenhausler2019causal}
D.~Rothenh{\"a}usler, P.~B{\"u}hlmann, and N.~Meinshausen, ``Causal dantzig: Fast inference in linear structural equation models with hidden variables under additive interventions,'' \emph{The Annals of Statistics}, vol.~47, no.~3, pp. 1688--1722, 2019.

\bibitem{RCA_Baselines_WWW20_AutoMAP}
M.~Ma, J.~Xu, Y.~Wang, P.~Chen, Z.~Zhang, and P.~Wang, ``Automap: Diagnose your microservice-based web applications automatically,'' in \emph{{WWW} '20: The Web Conference 2020}, 2020, pp. 246--258.

\bibitem{RCA_others_Seive_thalheim2017sieve}
J.~Thalheim, A.~Rodrigues, I.~E. Akkus, P.~Bhatotia, R.~Chen, B.~Viswanath, L.~Jiao, and C.~Fetzer, ``Sieve: Actionable insights from monitored metrics in distributed systems,'' in \emph{Proceedings of the 18th ACM/IFIP/USENIX Middleware Conference}, 2017, pp. 14--27.

\bibitem{spirtes2000causation}
P.~Spirtes, C.~N. Glymour, and R.~Scheines, \emph{Causation, prediction, and search}.\hskip 1em plus 0.5em minus 0.4em\relax MIT press, 2000.

\bibitem{entner2010causal}
D.~Entner and P.~O. Hoyer, ``On causal discovery from time series data using fci,'' \emph{Probabilistic graphical models}, pp. 121--128, 2010.

\bibitem{CD_hidden_AISTATS21_ABIC}
R.~Bhattacharya, T.~Nagarajan, D.~Malinsky, and I.~Shpitser, ``Differentiable causal discovery under unmeasured confounding,'' in \emph{The 24th International Conference on Artificial Intelligence and Statistics, {AISTATS} 2021}, ser. Proceedings of Machine Learning Research, vol. 130, 2021, pp. 2314--2322.

\bibitem{DBLP:conf/icdm/XiaZRGZ23}
Y.~Xia, H.~Zhang, Y.~Ren, J.~Guan, and S.~Zhou, ``Causal discovery by continuous optimization with conditional independence constraint: Methodology and performance,'' in \emph{{IEEE} International Conference on Data Mining, {ICDM} 2023}, 2023, pp. 668--677.

\bibitem{DBLP:conf/clear2/LiuHGKBG24}
W.~Liu, B.~Huang, E.~Gao, Q.~Ke, H.~D. Bondell, and M.~Gong, ``Causal discovery with mixed linear and nonlinear additive noise models: {A} scalable approach,'' in \emph{Causal Learning and Reasoning}, vol. 236, 2024, pp. 1237--1263.

\bibitem{CD_CDNOD_huang2020causal}
B.~Huang, K.~Zhang, J.~Zhang, J.~Ramsey, R.~Sanchez-Romero, C.~Glymour, and B.~Sch{\"o}lkopf, ``Causal discovery from heterogeneous/nonstationary data,'' \emph{The Journal of Machine Learning Research}, vol.~21, no.~1, pp. 3482--3534, 2020.

\bibitem{CD_intv_lippe2021efficient}
P.~Lippe, T.~Cohen, and E.~Gavves, ``Efficient neural causal discovery without acyclicity constraints,'' in \emph{International Conference on Learning Representations}, 2021.

\bibitem{pearl2009causality}
J.~Pearl, \emph{Causality}.\hskip 1em plus 0.5em minus 0.4em\relax Cambridge university press, 2009.

\bibitem{causal_bg_causalSurvey/tkdd/YaoCLLGZ21}
L.~Yao, Z.~Chu, S.~Li, Y.~Li, J.~Gao, and A.~Zhang, ``A survey on causal inference,'' \emph{{ACM} Trans. Knowl. Discov. Data}, vol.~15, no.~5, pp. 74:1--74:46, 2021.

\bibitem{CD_ADMG_Magnified_pena2016learning}
J.~M. Pe{\~n}a, ``Learning acyclic directed mixed graphs from observations and interventions,'' in \emph{Conference on Probabilistic Graphical Models}, 2016, pp. 392--402.

\bibitem{DBLP:conf/uai/AssaadDG22}
C.~K. Assaad, E.~Devijver, and {\'{E}}.~Gaussier, ``Discovery of extended summary graphs in time series,'' in \emph{Uncertainty in Artificial Intelligence, Proceedings of the Thirty-Eighth Conference on Uncertainty in Artificial Intelligence, {UAI} 2022}, J.~Cussens and K.~Zhang, Eds., vol. 180.\hskip 1em plus 0.5em minus 0.4em\relax {PMLR}, 2022, pp. 96--106.

\bibitem{Method_Related_CD_99_augLag}
A.~Nemirovsky, ``Optimization ii. numerical methods for nonlinear continuous optimization,'' 1999.

\bibitem{CL_survey_wang2021survey}
X.~Wang, Y.~Chen, and W.~Zhu, ``A survey on curriculum learning,'' \emph{IEEE Transactions on Pattern Analysis and Machine Intelligence}, vol.~44, no.~9, pp. 4555--4576, 2021.

\bibitem{Method_Related_CL_nipsworkshop23_dahmani2023child}
A.~Dahmani, E.~Yiu, T.~Lee, N.~Ke, O.~Kroemer, and A.~Gopnik, ``From child's play to ai: Insights into automated causal curriculum learning,'' in \emph{Intrinsically-Motivated and Open-Ended Learning Workshop@ NeurIPS2023}, 2023.

\bibitem{RCA_Others_ZhangKun22}
W.~Yang, K.~Zhang, and S.~Hoi, ``A causal approach to detecting multivariate time-series anomalies and root causes,'' 2022.

\bibitem{datasets_li2023causal}
P.~Li, Y.~Meng, X.~Wang, F.~Shen, Y.~Li, J.~Wang, and W.~Zhu, ``Causal discovery in temporal domain from interventional data,'' in \emph{Proceedings of the 32nd ACM International Conference on Information and Knowledge Management}, 2023, pp. 4074--4078.

\bibitem{datasets_mathur2016swat}
A.~P. Mathur and N.~O. Tippenhauer, ``Swat: A water treatment testbed for research and training on ics security,'' in \emph{2016 international workshop on cyber-physical systems for smart water networks (CySWater)}, 2016, pp. 31--36.

\bibitem{CD_KK/icml/LiuK23}
C.~Liu and K.~Kuang, ``Causal structure learning for latent intervened non-stationary data,'' in \emph{International Conference on Machine Learning, {ICML} 2023, 23-29 July 2023, Honolulu, Hawaii, {USA}}, ser. Proceedings of Machine Learning Research, vol. 202, 2023, pp. 21\,756--21\,777.

\bibitem{dataset_CD4swat_maiti2023iccps}
R.~R. Maiti, S.~Adepu, and E.~Lupu, ``Iccps: Impact discovery using causal inference for cyber attacks in cpss,'' \emph{arXiv preprint arXiv:2307.14161}, 2023.

\end{thebibliography}


\end{document}